\documentclass{bmvc2k}

\usepackage{amsmath}
\usepackage{algorithm}
\usepackage{algpseudocode}
\usepackage{graphicx}
\usepackage{tabularx}
\usepackage{amsfonts}       % blackboard math symbols
\usepackage{graphicx,wrapfig,lipsum}

\usepackage{subcaption}
\usepackage{xcolor}
\usepackage{multirow}
\usepackage{hhline}
\usepackage{xcolor} % For coloring text
\usepackage{enumitem}
\setlist[itemize]{topsep=0pt,itemsep=0pt, parsep=0pt,partopsep=0pt}
\title{Enabling Local Editing in Diffusion Models by Joint and Individual Component Analysis}
\addauthor{\small Theodoros Kouzelis}{ theodoros.kouzelis@athenarc.gr}{1,2}
\addauthor{\small Manos Plitsis}{ manos.plitsis@athenarc.gr}{3,4}
\addauthor{\small Mihalis A. Nicolaou}{m.nicolaou@cyi.ac.cy}{5}
\addauthor{\small Yannis Panagakis}{yannisp@di.uoa.gr}{2,4}

\addinstitution{
National Technical University of Athens,\\
Athens,Greece
}
\addinstitution{
 Archimedes AI, Athena RC,\\ Athens, Greece
}
\addinstitution{
 Institute for Language and Speech, Athena RC\\ Processing, Athens,Greece
}
\addinstitution{
 Department of Informatics and Telecommunications, National and Kapodistrian University of Athens,\\Athens,Greece
}
\addinstitution{
 Computation-based Science and Technology Research Center, The Cyprus Institute, \\ Nicosia, Cyprus
}
\runninghead{Kouzelis et al.}{Enabling local editing in Diffusion Models}

\begin{document}

\maketitle
\vspace{-12 pt}
\begin{abstract}
Recent advances in Diffusion Models (DMs) have led to significant progress in visual synthesis and editing tasks, establishing them as a strong competitor to Generative Adversarial Networks (GANs). However, the latent space of DMs is not as well understood as that of GANs. Recent research has focused on unsupervised semantic discovery in the latent space of DMs by leveraging the bottleneck layer of the denoising network, which has been shown to exhibit properties of a semantic latent space. However, these approaches are limited to discovering global attributes. In this paper we address, the challenge of local image manipulation in DMs and introduce an unsupervised method to factorize the latent semantics learned by the denoising network of pre-trained DMs. Given an arbitrary image and defined regions of interest, we utilize the Jacobian of the denoising network to establish a relation between the regions of interest and their corresponding subspaces in the latent space. Furthermore, we disentangle the {\it joint} and {\it individual} components of these subspaces to identify latent directions that enable {\it local} image manipulation. Once discovered, these directions can be applied to different images to produce semantically consistent edits,  making our method suitable for practical applications.
Experimental results on various datasets demonstrate that our method can produce semantic edits that are more localized and have better fidelity compared to the state-of-the-art. 
\href{https://zelaki.github.io/localdiff/}{https://zelaki.github.io/localdiff/}

%Advances in the understanding of Diffusion Models (DMs)
%have led to remarkable progress in visual editing and synthesis tasks and established them as a strong competitor to Generative Adversarial Networks (GANs). However, their latent space is still not as well understood. Recent works that focus on unsupervised semantic discovery in DMs are limited to finding global attributes. In this work, we address local image manipulation in DMs and present a method to discover the latent localized semantics learned by the denoising network of pre-trained DMs \textit{in an unsupervised manner}.
%Concretely, given an arbitrary image and some regions
%of interest, 
% we utilize the Jacobian of the denoising network to correlate specific regions of interest in an arbitrary image to corresponding latent subspaces and further decompose their joint and individual components, disentangling directions that capture global and local variation respectively. Experimental results on various datasets demonstrate the effectiveness of our method and its superiority over existing alternatives on regional fine-grained attribute control. 
\end{abstract}

\vspace{-13 pt}

\begin{figure}[H]
  \centering
  
    \includegraphics[width=9cm]{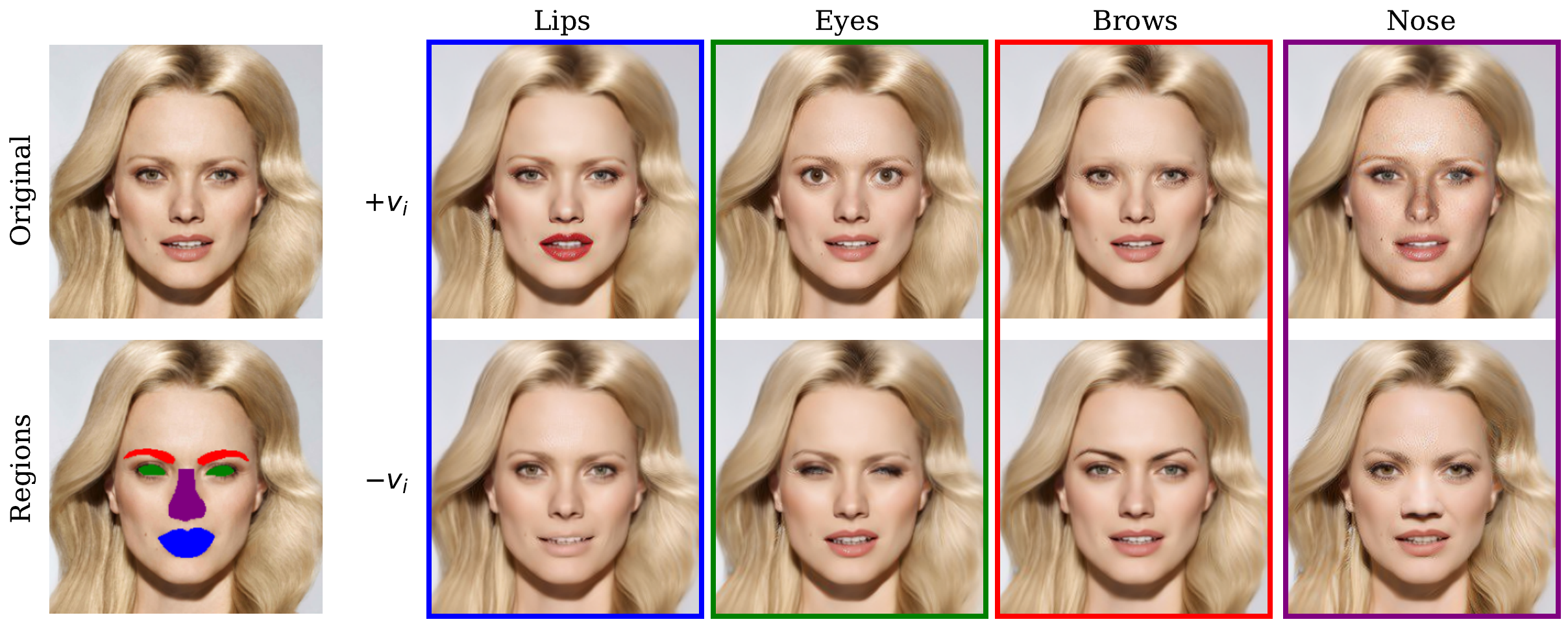}
    \vspace{7 pt}
  \caption{\textbf{Local Editing with our method}: Given regions of interest we can identify latent directions that result in diverse semantic edits without affecting the rest of the image. Linear interpolation within the identified semantic directions leads to gradual changes in the generated image like opening and closing the eyes. }
  \label{fig:intro}
\end{figure}

%------------------------------------------------------------------------- 
\section{Introduction}

Diffusion models \cite{ho2020denoising} have emerged as the new state-of-the-art paradigm of deep generative models.
They have surpassed the long-standing dominance of Generative Adversarial Networks (GANs) \cite{goodfellow2020generative} in image synthesis \cite{dhariwal2021diffusion} and they have also shown strong potential in a variety of computer vision tasks, such as text-guided image synthesis \cite{rombach2022high, dalle2, glide, imagen}, image editing \cite{hertz2022prompt, plug, imagic, diffedit} and inverse problems \cite{inverse1, inverse2}.

% However, while GANs by definition generate images by sampling from a semantic latent space, in diffusion models the definition of such a space is more challenging. 

% However, in contrast to GANs, their inner workings remain largely unexplored as the iterative generation process deems the definition of a latent space more challenging. The definition of latent space is essential for controlling the generative process and obtaining editing capabilities. 
\looseness-1However, while GANs generate images by sampling from a semantically informed latent space \cite{goodfellow2020generative}, that can then be used to guide the generative process and obtain editing capabilities \cite{harkonen2020ganspace, shen2020interpreting}, in DMs such a semantic latent space is harder to identify, mostly due to the iterative nature of the diffusion process.
Some approaches \cite{dhariwal2021diffusion, Avrahami2021BlendedDF, liu2023more, kim2022diffusionclip}  edit the latent variables (i.e., intermediate noisy images) guiding the generative process to a desired output, but require training a classifier \cite{dhariwal2021diffusion, Avrahami2021BlendedDF, liu2023more} or fine-tuning the whole model for each new attribute \cite{kim2022diffusionclip}.
Recently, Kwon et al. \cite{kwon2022diffusion} discovered that the bottleneck layer of the denoising U-net \cite{unet} (coined $\mathcal{H}$-space) possesses the properties of a semantic latent space. Building upon this finding, recent works have attempted to discover “interpretable directions” in the generator’s latent space \cite{haas2023discovering, park2023understanding}. Once discovered, these latent representations of high-level concepts can be utilized
to bring about predictable changes to the images. However, the latent directions discovered by previous works \cite{haas2023discovering, park2023understanding, kwon2022diffusion} result in global image manipulation without the ability of fine-grained regional control. 

In this paper, we develop, to the best of our knowledge, the first unsupervised approach for local image editing in unconditional DMs.
% by identifying subspaces in $\mathcal{H}$-space that capture local and global variation.
% we examine the latent space of unconditional DMs and manage to identify latent subspaces that reliably affect interpretable modes of variation explicitly within a region of interest. 
Given a pre-trained DM, a real image, and a set of regions of interest, such as the eyes or the lips in a face image (see Figure \ref{fig:intro}), our goal is to discover interpretable directions that specifically alter the selected image regions.  Firstly,  
% We then utilize the Singular Value Decomposition (SVD) of the Jacobian matrix of the generator which naturally yields latent subspaces spanned by directions that target specific regions of interest in an image. 
% Initially, we show that the row subspace induced by the Singular Value Decomposition (SVD)
% of the Jacobian of a region of the denoising network is spanned by directions that target specific regions of interest in an image. 
%In order to discover interpretable directions that specifically alter the selected image regions, 
we utilize the row space of the Jacobian constrained to each specified region, which is obtained by its Singular Value Decomposition (SVD). The row space is spanned by directions that manipulate the attributes in each region of interest. 
% we utilize the Singular Value Decomposition (SVD) of the Jacobian matrix of the generator, constrained to each specified region. Specifically, the row space of each Jacobian, spanned by its right singular vectors, contains directions that manipulate the attributes in each region of interest.
However,this approach lacks an explicit constraint ensuring localized edits, and hence other regions are inadvertently affected.
% Although such directions are guaranteed to maximize the variability in the region of interest, they also affect the rest of the image as they lack an explicit constraint ensuring localized edits.
% Although such directions result in desirable changes within the region of interest, they also inadvertently alter other parts of the image, as the directions identified by the Jacobian matrix are guaranteed to maximize the variability in the region of interest without an explicit constraint regarding the rest of the image.
% We manage to relate the latent space and the image region by computing the SVD of the Jacobian matrix of the denoising network,
% We then utilize the Singular Value Decomposition (SVD) of the Jacobian matrix of the generator which naturally yields latent subspaces spanned by directions that target specific regions of interest in an image. 
% We further observe that although such directions result in desirable changes within the region of interest they inadvertently alter other parts of the image. 
To alleviate this and achieve local manipulation, we further
propose decomposing the Jacobian associated with each region of interest into 
two distinct components: a \textit{joint} and an \textit{individual} component. 
% The joint component captures global modes of variation in the dataset, while the individual components, that are orthogonal to the joint, capture variation that 
% These are then used in place of the initial jacobians
The row space of the joint component comprises latent directions that induce global changes across the entire image. In contrast, the row space of the individual component, which is orthogonal to the joint, is spanned by latent directions that specifically target a designated region of interest without influencing other regions. 
To obtain this decomposition we utilize the so-called \textit{Joint and Individual Variation Explained (JIVE)} \cite{lock2013joint} method which is an iterative algorithm that estimates the joint and individual components in an arbitrary number of matrices. We further observe that the directions discovered from local regions of one image are readily applicable to other images, producing the same semantic manipulation, thus alleviating the need to recompute the decomposition for every sample.

Our main contributions can be summarized as follows:
\begin{itemize}

    \item We propose the first method that identifies semantic directions in the latent space of unconditional DMs that are localized to specific image regions, thus enabling local editing.%, enabling localized editing of specific image regions.
    %that edit a specific local image region.
    % \item We propose the first method for local editing within a region of interest with unconditional DMs. 
    % \vspace{-10 }
    % \textit{without} the need for additional training or external supervision.
    \item Local editing is achieved in an \textit{unsupervised manner} by decomposing the set of Jacobians that correspond to different image regions in joint and individual components capturing global and local variation respectively.
    \item We demonstrate that the semantic directions discovered by our method generalize from one image to others making them ideal for plug-and-play applications.
    \item We show both qualitatively and quantitatively the superiority of our approach for local image editing against existing alternatives, even supervised ones.
\end{itemize}

\section{Related Work}

\noindent \textbf{Diffusion Models} Diffusion Models \cite{sohl2015deep, ho2020denoising} continue to push forward the
state-of-the-art for image synthesis through architectural advances such as Latent Diffusion \cite{rombach2022high} and speeding up the generation process \cite{song2020denoising, song2023consistency}. Song et al. \cite{song2020score} have integrated DMs
and score-based models \cite{song2019generative} under an SDE formulation, improving our understanding of DMs as a reverse diffusion
process.  Classifier guidance \cite{dhariwal2021diffusion} and its variants \cite{sehwag2022generating, Avrahami2021BlendedDF, glide} control the generation process by guiding it toward a specific class. In \cite{preechakul2022diffusion} an additional encoder is introduced to capture semantic variation and control the generation process. However such approaches trade controllability with additional inference and training costs respectively.
Instead, Kwon et al. \cite{kwon2022diffusion} showed that the bottleneck layer of of-the-self DMs can be utilized to guide the generative process, exhibiting properties of a semantic latent space.

\noindent \looseness-1\textbf{Interpretable Latent Directions in Generative Models } Following the success of deep generative models in generating realistic and diverse images there has been a surge of interest in understanding the structure of their latent space. Many works \cite{harkonen2020ganspace, choi2021not, zhu2021low, zhu2022region,kwon2022diffusion, haas2023discovering,park2023understanding} aim to identify latent subspaces that capture meaningful semantic variation in the generated images. Most notably for GANs, \cite{harkonen2020ganspace} find such subspaces by applying Principal Component Analysis (PCA) to the intermediate generator’s representations. In \cite{choi2021not, choi2022finding} semantic latent subspaces are found by the SVD of the Jacobian matrix. Most related to our work, \cite{zhu2021low, zhu2022region} relate a latent subspace with a specific image region by leveraging the gradient of the GAN generator, while \cite{oldfield2022panda} operate directly on the feature maps and jointly discover factors representing spatial parts and their appearances.

In DMs, following the work of Kwon et al. \cite{kwon2022diffusion} recent works aim to discover interpretable directions in the latent space. In \cite{haas2023discovering}, they leverage the Jacobian of the generator to identify a semantic subspace in $\mathcal{H}$ without any supervision and \cite{park2023understanding} utilize the linearity of $\mathcal{H}$ to pull-back the metric tensor from $\mathcal{H}$ to the image space, establishing a semantic subspace. However, the latent directions detected by these methods tend to control
global image attributes whereas we disentangle the directions responsible for global and local edits.

\begin{figure}[t!]
  \centering
    \includegraphics[width=13cm]{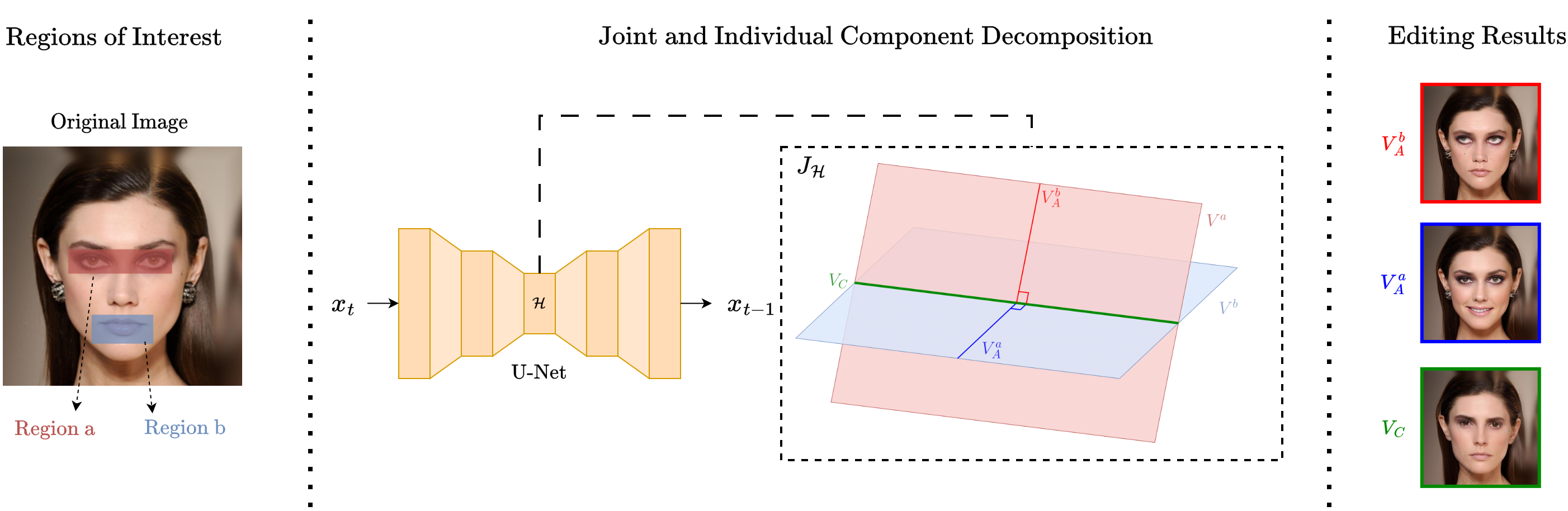}
    \vspace{2 pt}
  \caption{\textbf{An overview of our method.} Left: The regions of interest are selected. In this example, region $a$ and region $b$ correspond to the eyes and the mouth respectively.
  Center: The row space of the Jacobian of each region $\mathbf{V}^a$ and $\mathbf{V}^b$ is decomposed to the joint subspace $\mathbf{V}_C$ and the individual subspaces $\mathbf{V}_A^a$, $\mathbf{V}_A^b$. Right: Editing in $\mathcal{H}$ with directions from the joint subspace results in global edits, whereas editing with directions from the individual subspaces results in localized edits.}
  \label{mehtod}
\end{figure}
\section{Preliminary}

\subsection{Diffusion Models and $\mathcal{H}$-Space}
Diffusion Models are a class of generative
models where generation is modeled as a denoising process. A forward diffusion process adds increasing amounts of Gaussian noise to an image $\mathbf{x_0}$ in $\text{T}$ steps, and a learned
reverse process gradually removes the noise. The forward process is defined as:
% \vspace{-8pt}
\begin{equation}
    \label{forward}
    \mathbf{x}_t = \sqrt{a_t}\mathbf{x_0} + \sqrt{1-a_t}\mathbf{e}, \;\; \; \mathbf{e} \sim \mathcal{N}(\mathbf{0},\mathbf{I})
\end{equation}
% \vspace{-5pt}
where $a_t$ defines the noise schedule. DDIM \cite{song2020denoising} redefines (\ref{forward}) as a non-Markovian process and the approximate reverse process becomes: 
% \vspace{-8pt}
\begin{equation}
    \label{reverse}
    \mathbf{x}_{t-1} = \sqrt{a_{t-1}} \underbrace{ \bigg( \frac{\mathbf{x}_t - \sqrt{1-a_t} \; \mathbf{e}_t^\theta(\mathbf{x}_t)}{\sqrt{a_t}} \bigg)}_{\mathbf{P}_t} + \underbrace{ \sqrt{1-a_t -\sigma_t^2}}_{\mathbf{D}_t} \; \mathbf{e}_t^\theta (\mathbf{x}_t) + \sigma_t \mathbf{z}_t 
\end{equation}
% \vspace{-8pt}
\noindent where $\mathbf{z}_t \sim \mathcal{N}(\mathbf{0}, \mathbf{I})$ and $\sigma_t = \eta \sqrt{(1-a_{t-1}) / (1-a_t)} \sqrt{1-a_{t} / a_{t-1}}$. When 
$\eta = 0$, the process becomes deterministic and guarantees nearly perfect inversion.
% \subsection{Editing in $\mathcal{H}$-Space }
% \begin{wrapfigure}{r}{4.5cm}
% \includegraphics[width=4.5cm]{images/hspace.png}
% \caption{$\mathcal{H}$-space}\label{wrap-fig:1}
% \end{wrapfigure} 
Kwon et al. \cite{kwon2022diffusion} observed that the bottleneck feature maps of the denoising U-Net exhibit the properties of a semantic latent space. 
Given a pre-trained
denoising network $\mathbf{e}^\theta(\cdot)$  they show that a semantic latent direction $\mathbf{v} \in \mathcal{H}$ 
that modifies the latent code $\mathbf{h_t}$ for every timestep of the denoising process can cause
a desirable semantic change in the output image. Thus the denoising process of Eq. \ref{reverse} becomes:
\begin{equation}
    \mathbf{x}_{t-1} = \sqrt{a_{t-1}} \; \mathbf{P}_t \Big(\mathbf{e}^{\theta}_t (\mathbf{x}_t \vert \mathbf{h}_t+ \alpha \mathbf{v}) \Big) 
    + \mathbf{D}_t \Big(\mathbf{e}^{\theta}_t (\mathbf{x}_t \vert \mathbf{h}_t + \alpha \mathbf{v}) \Big)
    \label{eq:hspace}
\end{equation}
\noindent 
where $\mathbf{e}^{\theta}_t (\mathbf{x}_t \vert \mathbf{h}_t + \alpha \mathbf{v})$ denotes adding $\alpha \mathbf{v}$ to the feature maps $\mathbf{h}_t$,  $\alpha$ indicates the editing strength and $\mathbf{v}$ is
assumed to be a unit vector i.e. $\mathbf{v}^T \mathbf{v} = 1$. In this work, once we have discovered a latent direction $\mathbf{v}$ as presented in Section \ref{method} we use the editing process in Eq. \ref{eq:hspace} to edit a region of interest.

\section{Methodology}
\label{method}
In this section, we describe our method in detail. In Section \ref{jacobian} we describe how the SVD of the Jacobian can identify a subspace spanned by directions that control the principal modes of variation in a region of interest. In Section \ref{jive}, given M regions of interest, we proceed to decompose the Jacobian of each region to a \textit{joint} and \textit{individual} component to achieve localized edits.

\subsection{Jacobian Decomposition}
\label{jacobian}

Let $\{\mathbf{e}^\theta_t (\mathbf{x}_t \vert \mathbf{h}_t)\}_m \in \mathbb{R}^{d_m}$ be the output of the denoising network in a specified region $\mathbf{m}$ where $d_m$ is the number of pixels in the region and the bottleneck $\mathbf{h}_t \in \mathbb{R}^{d_h}$ is of dimension $d_h$. Then the derivative of the output region w.r.t. $\mathbf{h}_t$ at timestep $t$ is given by the Jacobian matrix \{$\mathbf{J}^t_h\}_m = \frac{\partial \{\mathbf{e}^\theta_t (\mathbf{x}_t \vert \mathbf{h}_t)\}_m}{ \partial \mathbf{h}_t} \in \mathbb{R}^{{d_m} \times {d_h}}$. This is a matrix whose rows are the derivatives of each pixel value within region $\mathbf{m}$ w.r.t. $\mathbf{h}_t$ and $d_m$ is the number of pixels in the region. For presentation purposes, we will refer to the Jacobian of an output region $\mathbf{m}$ as $\mathbf{J}^{(m)}$ for the rest of the paper.
Given an arbitrary vector $\mathbf{v} \in \mathbb{R}^{d_h}$,
the directional derivative:
\begin{equation}
    \underset{\epsilon \rightarrow 0}{\text{lim}} \frac{ \{ \mathbf{e}^\theta_t (\mathbf{x}_t \vert \mathbf{h}_t + \epsilon \mathbf{v}) \}_m
    - \{ \mathbf{e}^\theta_t (\mathbf{x}_t \vert \mathbf{h}_t) \}_m}{\epsilon}  =\mathbf{J}^{(m)} \mathbf{v} 
\end{equation}
\noindent measures the instantaneous change in $\{\mathbf{e}^\theta_t \}_m$ resulting from a
perturbation of $\mathbf{h}_t$ along the direction of $\mathbf{v}$. The unit-norm
perturbation of $\mathbf{h}_t$ that maximizes the magnitude of this change is $\mathbf{v}_1 := \underset{\mathbf{v}}{\text{argmax}} \Vert \mathbf{J}^{(m)} \mathbf{v} \Vert$. This is the first right singular vector of $\mathbf{J}^{(m)}$. Hence, a perturbation of $\mathbf{h}_t$ along $\mathbf{v}_1$ maximizes
the magnitude of the instantaneous change in the output noisy image at timestep $t$. By maximizing $\Vert \mathbf{J}^{(m)} \mathbf{v} \Vert$ while remaining
orthogonal to $\mathbf{v}_1$, one can derive the second right singular vector $\mathbf{v}_2$.  By continuing this process we obtain $r$ directions in $\mathcal{H}$-space that maximize the variability of the noisy image at time t.
Thus the right singular vectors $\mathbf{V}^{(m)}$ from the SVD of the Jacobian i.e. $\mathbf{J}^{(m)} = \mathbf{U}^{(m)} \mathbf{S}^{(m)} \mathbf{V}^{{(m)}^T}$ span a subspace i.e. the row space of the Jacobian, that captures the principal modes of variation in the region of interest.

\looseness-1In practice it is highly inefficient to estimate the Jacobian of the denoising U-net directly, thus we rely on the \textit{subspace iteration} method \cite{haas2023discovering} to approximate the SVD of the $\mathbf{J}^{(m)}$ without ever storing it to memory.  For a detailed description of the algorithm, please refer to \cite{haas2023discovering} and the Appendix.

\subsection{Joint and Individual Components in the Latent Space of DMs}
\label{jive}

The method described above lacks an explicit constraint that ensures localized edits.
Note that a latent vector that maximizes the variability in a specified region can inadvertently affect other regions. However,
for local editing, we would like to manipulate
a specified region while not affecting the rest of the image.
Our idea is to \textit{disentagle} the \textit{joint} and \textit{individual} components of the Jacobian of each region. In this manner editing within the row space of the joint component results in global edits whereas directions from the row space of the individual component result in local edits.

Formally, given a real image $\mathbf{I}$, a set of $N$ image regions that segment the image into parts i.e. $M = [\; \mathbf{m}_i \; \vert \; i \in (1,\; \cdots, \; N), \; \cup_{i=1}^N \{\mathbf{m}_i \}= \mathbf{I} \;]$ and the Jacobian of each region  $\{ \mathbf{J}^{(i)} \}_{i=1}^N$  we seek to decompose each Jacobian to a joint and an individual component \mbox{$\mathbf{J}^{(i)} \approx \mathbf{C}^{(i)} + \mathbf{A}^{(i)}$} that adheres to the following properties:
\begin{itemize}
    \item  The row spaces of the matrices capturing joint variation, i.e., joint matrices $\mathbf{C}^{(i)}$, are defined as sharing a common subspace denoted as $\text{Row}(\mathbf{C}) = \text{Row}(\mathbf{C}^{(i)}), \; \forall i \in (1, \; \cdots \; , N)$
    \item Components $\mathbf{A}^{(i)}$ are deemed individual since they are imposed to be orthogonal to the joint component, i.e. $\text{Row}(\mathbf{C}) \perp \text{Row}(\mathbf{A}^{(i)}), \; \forall i \in (1, \; \cdots \; , N)$

    \item The intersection of the row subspaces of the individual components is the zero vector space, \mbox{$\cap_{i=1}^N \text{Row}(\mathbf{A}_i) = \mathbf{0}$}
\end{itemize}

\noindent Let $\mathbf{J} = \Big[\mathbf{J}^{(1)^T}, \; \cdots \;, \mathbf{J}^{(N)^T} \Big]^T \in \mathbb{R}^{q \times d_h}$ be the concatenation of the Jacobians of each region along their rows, where $q = d_m^{(1)} +  \; \cdots \; + d_m^{(N)}$. The joint and individual components are obtained by solving the following constrained
optimization problem:

\begin{align}
    &\underset{C, \{ A^{(i)}\}_{i=1}^N}{\text{min}} \;  \Big\Vert \mathbf{J} - \mathbf{C} - 
    \big[ \mathbf{A}^{(1)^T}, \; \cdots \;, \mathbf{A}^{(n)^T} \big]^T
    \Big\Vert_{\text{F}}^2  \\
    \text{s.t.} \;\;\; &\text{rank}(\mathbf{C})=r_C, \;\; 
    \{ \text{rank}(\mathbf{A}^{(i)}) = r_A^{(i)}, \mathbf{C}\mathbf{A}^{(i)^T}=\mathbf{0} \}_{i=1}^N, \nonumber
\end{align}

\noindent where  $\mathbf{C} = \Big[\mathbf{C}^{(1)^T}, \; \cdots \;, \mathbf{C}^{(N)^T} \Big]^T \in \mathbb{R}^{q \times d_h}$ and $\{\mathbf{A}^{(i)} \in \mathbb{R}^{d_m^{(i)} \times d_h}\}_{i=0}^N$ are the joint and individual components of $\mathbf{J}$ respectively. We approximate the solution of this optimization problem by utilizing the iterative JIVE method as proposed in \cite{lock2013joint}.

% \noindent We use the JIVE method to approximate this decomposition. We find that a large joint and a small individual rank $r_C < r_{A_i}$ yield the best results for local editing. This aligns with our intuition since the global modes of variation are expected to be more than the local.

After the decomposition, the row subspace $\mathbf{V}_C$ obtained by the SVD of the joint component $\mathbf{C}=\mathbf{U}_C \mathbf{S}_C \mathbf{V}_C$ captures global variation in the entire image (e.g. changing the gender). In contrast, the row subspace $\mathbf{V}^{(i)}_A$ obtained by the SVD of the individual component $\mathbf{A}^{(i)} =  \mathbf{U}^{(i)}_A \mathbf{S}^{(i)}_A \mathbf{V}^{{(i)}^T}_A$ captures local modes of variation specific to region $\mathbf{m}^{(i)}$. Figure \ref{mehtod} (middle) illustrates the effect of JIVE on the row subspaces of the Jacobians.

Calculating the JIVE decomposition directly on the set of Jacobians $\mathbf{J} \in \mathbb{R}^{q\times d_h}$ is highly impractical. For instance, in a standard DDPM  \cite{ho2020denoising}, with image size $q=256 \cdot 256 \cdot 3$ and latent dimension $d_h = 8 \cdot 8 \cdot 512$ the Jacobian 
 $\mathbf{J}$ has approximately $6\text{B}$ parameters. 
 The most computationally expensive step in the JIVE algorithm (see \cite{lock2013joint}) is that of SVD. Clearly applying SVD on a 6B parameter Jacobian is prohibitive in practice. To make such a computation feasible we adopt a dimension-reducing transformation: $\mathbf{J}^{(i)}  \rightarrow \mathbf{J}^{(i)}_\perp$
where $\mathbf{J}^{(i)}_\perp =  \mathbf{S}^{(i)}  \mathbf{V}^{{(i)}^T}$ is an $r \times  d_h$ matrix with $r \ll q$, derived from the SVD of $\mathbf{J}^{(i)}$ as obtained from the \textit{subspace iteration} \ref{jacobian}. The above mentioned approach is valid since the Euclidian distance between the columns of $\mathbf{J}^{(i)} $ is
preserved in $\mathbf{J}^{(i)}_\perp$ \cite{lock2013joint}.

\section{Experiments}

In this section, we present a series of experiments to validate the proposed method.
Initially, we describe our experimental setup in Section \ref{setup}.
Then, in Section \ref{exp_indiv} we showcase the effectiveness of the individual components on localized edits.
In Section \ref{qual_res} we show that our method can identify meaningful editing directions from a single image that generalize to other images. Finally in Section \ref{compare} we qualitatively and quantitatively compare our approach with existing alternatives for attribute manipulation in unconditional DMs.

\subsection{Experimental Setup}
\label{setup}
We conduct our experiments on three different datasets, namely CelebA-HQ \cite{liu2015faceattributes}, LSUN-churches \cite{yu15lsun}, and M\textsc{et}F\textsc{aces} \cite{karras2020training}, using an unconditional DDPM\footnote{https://huggingface.co/google/ddpm-ema-celebahq-256} \footnote{https://huggingface.co/google/ddpm-ema-church-256} \footnote{https://github.com/jychoi118/P2-weighting} as the base model. We highlight that all models are pre-trained and kept frozen. 
 We find that a large joint and a small individual rank $r_C < r_{A_i}$ yield the best results for local editing. This aligns with our intuition since the global modes of variation are expected to be more than the local.
For all experiments presented, we set the rank of our dimension-reducing transformation to $r=50$, the joint and individual rank to $r_C=30$ and $r_A = 5$ respectively, and obtain the editing directions at timestep $t=0.6T$. Editing results derived from different timesteps can be found in the Appendix.

% \begin{figure}[H]

%   \centering
%     \includegraphics[width=0.9\textwidth]{images/main.pdf}
%     \vspace{5 pt}
%   \caption{ \textbf{Local editing results} on the Celeba-HQ. The region of interest is highlighted with \textcolor{magenta}{pink rectangles}.  Our method can identify diverse semantic manipulations within a region while not affecting the rest of the image. Note that the latent vectors used to edit the images in each row are derived from the image in the first row.}
% \label{fig:celeba}

% \end{figure}

% \begin{figure}[h]

%   \centering
%     \includegraphics[width=0.9\textwidth]{images/main_other.pdf}
%     \vspace{5 pt}

%   \caption{Individual subspace editing for LSUN churches and bedrooms.}
% \label{fig:lsun_indiv}
% \end{figure}

For quantitative evaluation, we use Fréchet Inception Distance (FID), Identity Similarity (ID), and Region of Interest Ratio (ROIR) \cite{oldfield2022panda}. FID is utilized to evaluate the fidelity of the generated images after the edit. To assess  identity similarity (ID) before and after the edit we use the ArcFace model \cite{Deng_2019_CVPR}. To quantify local editing, we use ROIR \cite{oldfield2022panda}, which is the ratio of the distance between pixels of the original and edited images in the
region of ‘disinterest’, over the same quantity in the region of interest.
A small ROIR indicates localized edits, with large changes within the region of interest and small changes in the rest of the image:
\begin{equation}
    \text{ROIR}(\mathcal{M}, \mathcal{X}, \mathcal{X}') = \frac{1}{N} \sum_{i=1}^N 
    \frac{\Vert (\mathbf{1} - \mathcal{M}) \cdot (\mathcal{X}_i - \mathcal{X}'_i)\Vert}
    {\Vert \mathcal{M} \cdot (\mathcal{X}_i - \mathcal{X}'_i)\Vert}
\end{equation}

\noindent where $\mathcal{M} \in [0,1]^{H\times W \times C}$ is the mask specifying the region of interest, $\mathbf{1}$ is a $1$-tensor and $\mathcal{X}, \mathcal{X}' \in \mathbb{R}^{N \times H \times W \times C}$ are a batch of original and edited images respectivly.

\begin{figure}[t!]

  \centering
    \includegraphics[width=\textwidth]{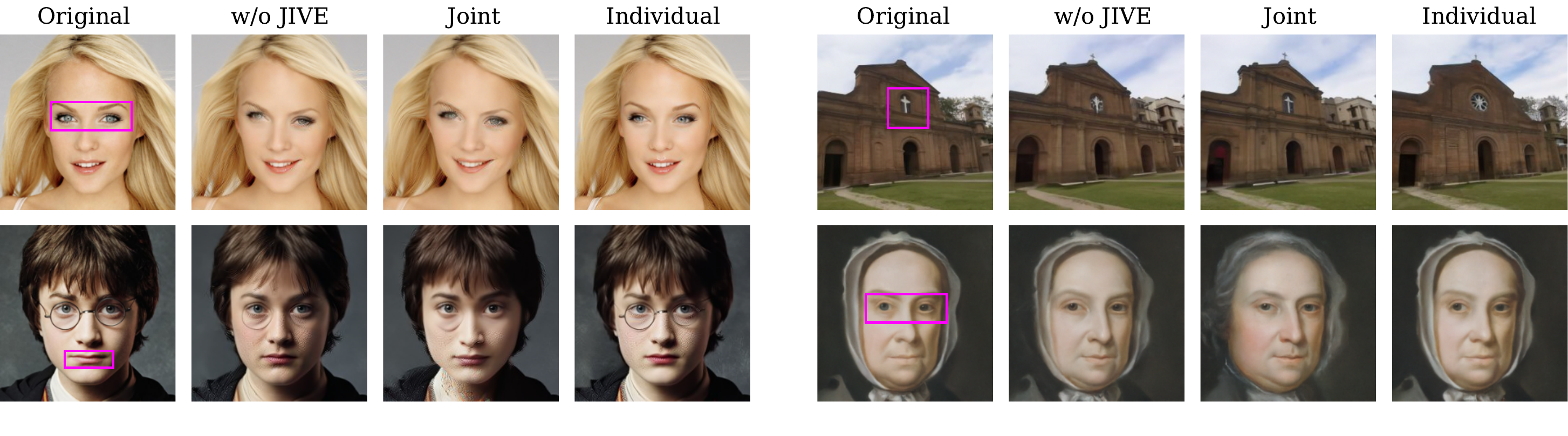}

  \caption{\textbf{Editing with the joint and individual components} for the CelebA-HQ, LSUN-Churches and M\textsc{et}F\textsc{aces} datasets. Regions of interest are denoted by a pink rectangle. By decomposing the Jacobians of each region into a joint and individual component we can disentangle the global and the local semantic variation}
\label{fig:jive_nojive}

\end{figure}

\subsection{Editing within the Individual and Joint Subspaces}
\label{exp_indiv}

\begin{figure}[t!] % 'h' for "here", you can use other placement specifiers
  \centering
  
  % First subfigure
  \begin{subfigure}[b]{\linewidth}
    \centering
    \includegraphics[width=0.9\linewidth]{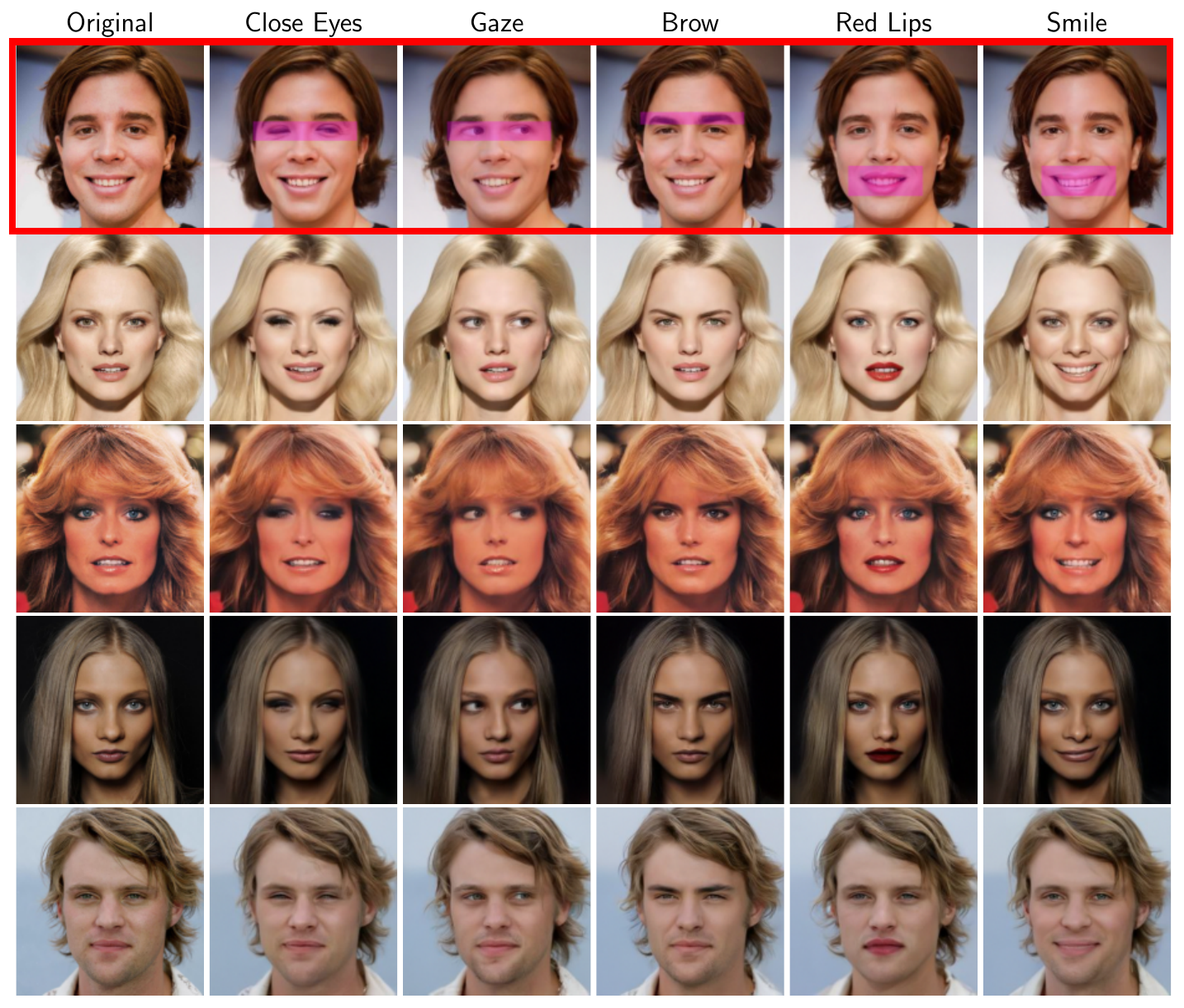}
    % \caption{CelebA-HQ}
    \label{fig:sub1}
  \end{subfigure}
  
  % Add some vertical space between the images

  \vspace{5pt}
  % Second subfigure
  \begin{subfigure}[b]{\linewidth}
    \centering
    \includegraphics[width=0.9\linewidth]{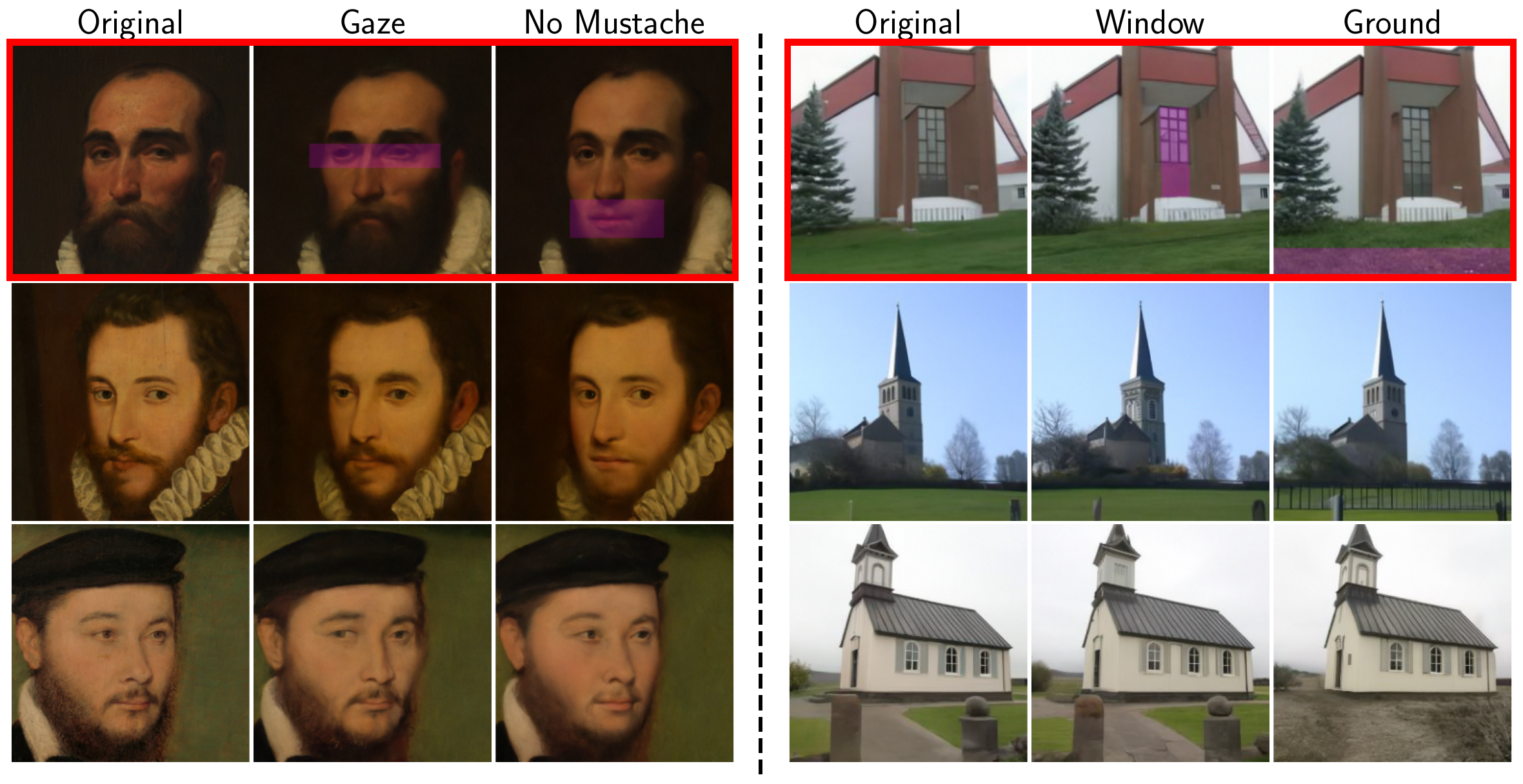}
    % \caption{Metfaces \& LSUN-Churches}
    \label{fig:sub2}
  \end{subfigure}
    \vspace{5pt}
  \caption{\textbf{Local editing results} on the CelebA-HQ (top), M\textsc{et}F\textsc{aces} and LSUN-Churches (bottom). The region of interest is highlighted with \textcolor{magenta}{pink rectangles}.  Our method can identify diverse semantic manipulations within a region while not affecting the rest of the image. Note that the latent vectors used to edit the images in each row are derived from the image in the first row.}
  \label{fig:test}
\end{figure}
\vspace{-5pt}

In this section, we validate the effectiveness of the JIVE decomposition on the Jacobians as presented in Section \ref{jive} on samples from CelebA-HQ, LSUN-Churches, and M\textsc{et}F\textsc{aces}. In Figure \ref{fig:jive_nojive} we show the effects of editing with latent vectors belonging to the row spaces of the Jacobian w/o JIVE, the individual component, and the joint component of the Jacobian respectively.  First, we observe that when we edit a region $\mathbf{m}_i$ directly within the row space of the Jacobian i.e. $\mathbf{v}_i \in \mathbf{V}^{(i)}$, undesirable non-localized edits occur. 
For example, when editing the lips of Harry Potter, other attributes such as the glasses and the color of the cheeks are manipulated and when editing the church window, the background building is also altered. In the next column, we depict the effects of editing within the row space of the joint component $\mathbf{v}_i \in \mathbf{V}_C^{(i)}$. This results in global manipulations that affect the entire image, such as editing both eyes and mouth in the top left image from CelebA-HQ and changing the gender in the sample from M\textsc{et}F\textsc{aces}.  On the contrary, when using the vectors from the row space of the individual component $\mathbf{v}_i \in \mathbf{V}_{A}^{(i)}$ to edit the image region the manipulations are highly localized %and the rest of the image is minimally affected
. In the last two rows of Table \ref{tab:results} we demonstrate this quantitatively for four attribute manipulations. When edits are derived from the individual component, FID and ROIR  decrease, while ID increases indicating that we could achieve more
precise control over a specific region, while better retaining image quality and identity similarity.

% In the top right image, when changing the mouth to a smile, not using JIVE or editing with the joint component visibly changes the eye position. In the top right, when changing the cross-shaped window, not using JIVE or editing with the Joint component affects the background buildings and the lighting inside the doors. In the bottom right picture, using the Individual component only changes the black pillows, whereas in the other two columns the whole bed is changed into two single beds.

\subsection{Qualitative results}
\label{qual_res}

Here we present qualitative results, as depicted in Fig. \ref{fig:test}. We show that directions obtained from the individual component extracted from a region of interest by our method can perform various localized semantic edits. We highlight with a red rectangle, how semantic directions identified on a single reference image are transferred to the rest for each dataset.
% As shown in Fig. \ref{fig:test}, the reference images
% and the corresponding pink masks indicate the images and the regions we use to find the attribute vectors, which are used to edit all the images. 
For CelebA-HQ samples, the eyes, eyebrows, and mouth regions are selected. For the eyes region, local semantic changes such as closing the eyes and changing the gaze are depicted in the second and third columns. In the fourth column, the eyebrows become thicker. In the last two columns where the mouth region is selected, we can manipulate the facial expression by adding a smile and changing the lip color. Similarly for M\textsc{et}F\textsc{aces}, we present directions that manipulate the gaze in the eyes region and remove the mustache in the mouth region. Finally for LSUN-Churches, directions that alter the windows and the ground are presented. 

% Note that regions of interest between the reference
% image and target images are not necessarily aligned (e.g. the window regions in LSUN-Churches). This suggests that the identified subspaces are semantic-aware.

% Figure \ref{fig:lsun_indiv} shows the same localized editing but for the LSUN bedrooms and churches datasets. We see that even with datasets containing images of buildings or interiors with different views and content, our method still manages to find meaningful local semantic directions for editing only e.g. the ground in front of a church, or the church tower windows.

% \begin{figure}[h]

%   \centering
%     \includegraphics[width=0.9\textwidth]{images/main_other.pdf}
%     \vspace{5 pt}

%   \caption{Individual subspace editing for LSUN churches and bedrooms.}
% \label{fig:lsun_indiv}
% \end{figure}

\subsection{Comparison with Other Methods}
\label{compare}

\begin{figure}[t!]

  \centering
    \includegraphics[width=0.9\textwidth]{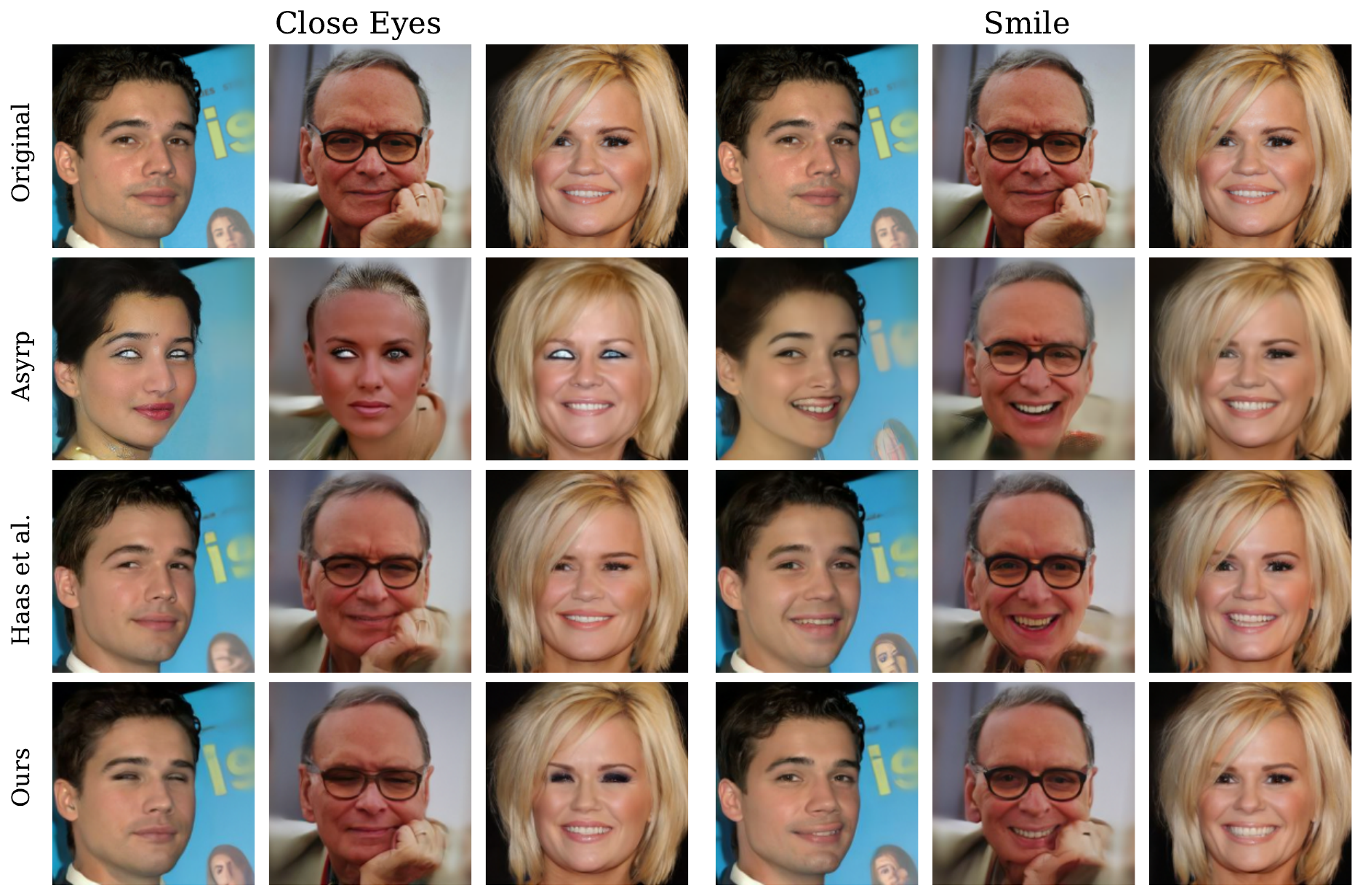}
    \vspace{5 pt}

  \caption{\textbf{Qualitative comparison} between our method and existing alternatives for two local edits.}
\label{fig:compare}
\end{figure}

% \vspace{-25pt}

% \usepackage{multirow}
\begin{table}[t!]
\centering
\scalebox{0.67}{
\begin{tabular}{|c|ccc|ccl|ccc|ccc|}
\hline
\multirow{2}{*}{\textbf{Method}} & \multicolumn{3}{c|}{\textbf{Close Eyes}}   & \multicolumn{3}{c|}{\textbf{Smile}}        & \multicolumn{3}{c|}{\textbf{Red Lips}}     & \multicolumn{3}{c|}{\textbf{Gaze}}         \\ \cline{2-13} 
                                 & \textbf{FID} $\downarrow$ & \textbf{ID} $\uparrow$ & \textbf{ROIR} $\downarrow$ & \textbf{FID} $\downarrow$ & \textbf{ID} $\uparrow$ & \textbf{ROIR} $\downarrow$ & \textbf{FID} $\downarrow$ & \textbf{ID} $\uparrow$ & \textbf{ROIR} $\downarrow$& \textbf{FID} $\downarrow$ & \textbf{ID} $\uparrow$ & \textbf{ROIR} $\downarrow$\\ \hline
Asyrp \cite{kwon2022diffusion}                          & 60.63        & 0.40        & 3.95          & 67.9         & 0.57        & 4.32          & 78.27        & 0.32        & 3.75          & 84.17        & 0.12        & 6.04          \\ \hline
Haas et al. \cite{haas2023discovering}                           & 51.42        & \textbf{0.70}        & 5.18          & 52.16        & 0.68        & 4.30          & \textbf{48.82}        & 0.69        & 6.61          & 52.28        & 0.68        & 4.05          \\ \hline
Ours (wo/ JIVE)                  & 52.53        & 0.67        & 3.66          & 54.20        & 0.72        & 3.62          & 51.46        & 0.73        & 4.32          & 51.23        & 0.68        & 3.53          \\ \hline
Ours (w/ JIVE)                   & \textbf{49.16}        & 0.69        & \textbf{2.87}          & \textbf{51.90}        & \textbf{0.78}        & \textbf{2.73}          & 48.93        & \textbf{0.74}        & \textbf{3.26}          & \textbf{48.11}        & \textbf{0.71}        & \textbf{3.07}          \\ \hline
\end{tabular}
}
\vspace{10pt}
\caption{\textbf{Quantitative results} between our method with and without the JIVE decomposition
and existing alternatives for four localized edits on 5k CelebA-HQ samples.}
\label{tab:results}
\end{table}

In this section, we compare our method both qualitatively and quantitatively with the state-of-the-art for attribute manipulation in unconditional DMs. Specifically, we 
compare our approach with Asyrp \cite{kwon2022diffusion}, and the method proposed by Haas et al. \cite{haas2023discovering}, two recently proposed methods that identify semantic directions in the latent space of unconditional DMs. We find the most relevant vectors that can control the eyes and smile according to their papers. Note that Asyrp is a supervised method, that uses CLIP \cite{radford2021learning} to achieve image edits. Also, the method of Haas et al. is equivalent to our method without using JIVE or constraining the Jacobians to a region of interest.
% We used the above methods as well as ours, to extract directions in H-space that affect the region of the eyes and mouth in three different images from CelebA, and chose the ones that close the eyes and produce a smile. 
% In order to be thorough in our comparisons we attempted to include Xspace \cite{park2023understanding} in the methods that we compare to, but we were unable to find directions that produce semantic changes in faces like closing the eyes or smiling.
As shown in Figure \ref{fig:compare}, our method retains the individual characteristics of the original image and the edits are better restricted to the region of interest than the other two methods. For the Close Eyes edit, Asyrp produces unrealistic artifacts in the eye region and even alters the gender of the first two images. While Haas et al. better retain the subject's identity, their method fails to fully close the eyes. For the Smile edit, Asyrp swaps the first image subject's gender, and also significantly alters the second image, removing the hand. Haas et al. fail to produce edits as localized as our method, changing the subject's facial expression and characteristics outside of the specified region.

To quantify the comparison, we present in Table \ref{tab:results} quantitative experiments for four different manipulations on 5k real images from CelebA-HQ. Our method achieves comparable ID and FID to Haas et al. for the Close Eyes and Red Lips edits respectively, while outperforming both methods on all metrics for the rest of the attributes. As captured by the FID and ID metrics, our method produces edited images of higher fidelity, more closely resembling the original images. Finally, as captured by the ROIR metric, our method is better at producing localized edits that do not affect the rest of the image.

% Even without using JIVE, by taking the Jacobian for specific regions of interest helps in localizing the resulting edits compared to existing methods, even when transferred to other images in the dataset.

\section{Conclusion and Future Work}

In this work, we propose a method for localized semantic manipulation of real images using a pre-trained DM. Our method involves first associating specific regions of interest in an image to subspaces in the DMs latent space and then factorizing these subspaces to isolate their individual and joint variation. We find that the subspaces discovered from one image can be used to edit different images, making the computation of a new factorization unnecessary. By extensive qualitative and quantitative experiments, we establish that our method can produce meaningful edits that are localized to specific regions of interest while preserving the original image quality and identity better than previous methods. In future work we aim to explore the latent semantics of video diffusion models. The temporal dimension in video generation adds complexity to the latent space, making it an interesting research direction to identify latent subspaces corresponding to specific temporal moments.

\vspace{20pt}

\noindent \textbf{Acknowledgement} This work has been partially supported by project MIS 5154714 of the Natonal Recovery and Resilience Plan Greece 2.0 funded by the European Union under the NextGeneraton EU Program and by a grant from The Cyprus Institute on Cyclone.

% \bibliography{egbib}
\clearpage

\appendix

 \newpage%
 \renewcommand{\thesection}{\Alph{section}}% For Alpha numeric number

 {\huge \textbf{Appendices}}

 \vspace{10pt}

\noindent In this document, we present additional material to support the main paper. Firstly, 
in Section A for reproducibility purposes 
we provide implementation details and detailed algorithms for our method. 
Lastly, in Section B we provide additional experimental results,
comparisons and an ablation study designed to supplement and further validate our proposed method.

\section{Algorithms \& Implementation Details}

\vspace{5 pt}

\subsection{Algorithms}
In this Section, we provide an algorithm that describes our method as presented in Section 4.2 in the main paper. Additionally, we offer detailed algorithms for the Subspace Iteration method used to approximate the SVD of the Jacobian of the denoising network and the JIVE algorithm used to obtain a solution to the minimization problem in Eq. (5) of the main paper.

\noindent \textbf{Computing Joint and Individual Components in the Latent Space of DMs}: Algorithm \ref{algo:main} summarizes our method for decomposing the Jacobian of each region into a joint and an individual component. We start with a real image $\mathbf{x}_0$, a denoising network $\mathbf{e}_{\theta}$ and a set of regions $M=\{ \mathbf{m}_1, \cdots, \mathbf{m}_N \}$, along with the denoising timestep $t$ and joint and individual ranks $r_C$ and $r_A$. Starting with $\mathbf{x}_0$, we obtain its corresponding noise  $\mathbf{x}_T$ via DDIM Inversion \cite{song2020denoising}. We denoise $\mathbf{x}_T$ up to timestep $t$ with the standard DDIM reverse process (see Eq. (3) in text). Then we obtain the SVD of the Jacobian of each region with the Subspace Iteration (Algorithm  \ref{algo:subsapce}). Finally, by utilizing the JIVE  (Algorithm \ref{algo:jive}) we obtain the joint and individual components $\mathbf{C}$ and $\mathbf{A}_i$ for each region $i$. The rows of $\mathbf{A}_i$ are latent directions that result in meaningful local edits within region $i$.

\begin{algorithm}[H]

\caption{Computing Joint and Individual Components in the Latent Space of DMs}
\label{algo:main}
\begin{algorithmic}[1]
\Procedure{ComputeIndivJoint}{$\mathbf{e}_\theta,\mathbf{x}_0,t , \text{M}, r_C, r_A$} 
 \State $\mathbf{x}_T \gets \text{DDIMInversion}(\mathbf{x}_0)$ \Comment{DDIM Inversion \cite{song2020denoising}}

  \State $\mathbf{x}_t \gets \text{DDIMReverse}(\mathbf{x}_T, t)$\Comment{DDIM Reverse process until timestep t}
 % \State $\mathbf{x_t \gets \text{DDIMReverse}(\mathbf{x}_T)$

\For{$i, \mathbf{m} \; \; \text{in} \; \; \text{enumerate}(\text{M})$}

    $\mathbf{U}^{(i)} , \mathbf{S}^{(i)} , \mathbf{V}^{(i)} \gets \text{SubspaceIteration}\big(\{ \mathbf{e}_\theta^t (\mathbf{x}_t) \}_m \big)$ \Comment{Algorithm \ref{algo:subsapce}}
    
    $\mathbf{J}_\perp^{(i)} \gets \mathbf{S}^{(i)} \mathbf{V}^{(i)^T}$ \Comment{Dimension-reducing transformation} 
    
\EndFor
  \State $\mathbf{J}_\perp = \big[  \mathbf{J}^{(1)^T} , \cdots, \mathbf{J}^{(N)^T} \big]^T$

    \State $\mathbf{C}, \{\mathbf{A}_i \}_{i=1}^N \gets \text{JIVE}(\mathbf{J}_\perp, r_C, r_A)$ \Comment{Algorithm  \ref{algo:jive}}
\State \textbf{return} $\mathbf{C}, \{\mathbf{A}_i \}_{i=1}^N$
\EndProcedure
\end{algorithmic}
\end{algorithm}

\noindent \textbf{Jacobian Subspace Iteration}: As 
 discussed in Section 4.2 the dimension of the latent space and the output image, result in a Jacobian matrix of approximately 6B parameters. To efficiently calculate the SVD of the Jacobian of the denoising network without storing it in memory we rely on Jacobian Subspace Iteration as proposed by Haas et al. \cite{haas2023discovering}. Note that differently from \cite{haas2023discovering} we calculate the SVD of the Jacobian of the denoising network explicitly within a region of interest $\mathbf{m}$.

\begin{algorithm}[H]
\caption{Jacobian Subspace Iteration}
\label{algo:subsapce}
\begin{algorithmic}[1]
\Procedure{SubspaceIteration}{$\mathbf{e}_\theta,\mathbf{x}_t,\mathbf{m}$} 
 \State $\mathbf{h}_t \gets \mathbf{e}_\theta^t (\mathbf{x}_t)_{:\frac{l}{2}}$\Comment{Apply only the encoder to get bottleneck featuremaps}

 \State $\mathbf{y}_m \gets \{ \mathbf{e}_\theta^t (\mathbf{x}_t) \}_m$\Comment{Output of denoising network in region $\mathbf{m}$}

 \State $\mathbf{V} \gets \text{i.i.d standard Gaussian Samples}$
 \State $\mathbf{Q}, \mathbf{R}  \gets \text{QR}(\mathbf{V})$ \Comment{Reduced QR decomposition}
 \State $\mathbf{V} \gets \mathbf{Q}$ \Comment{s.t. $\mathbf{V} \mathbf{V}^T = \mathbf{I}$}

 \While{termination criterion}
 \vspace{5pt}

   $ \mathbf{U} \gets \frac{\partial \{ \mathbf{e}_\theta^t (\mathbf{x}_t \vert \ \mathbf{h}_t + a \mathbf{V}) \}_m }{\partial a} \Big \vert_{a=0}$ \Comment{Forward differentiation} 
   
   $\hat{\mathbf{V}} \gets \frac{\partial \mathbf{U}^T \mathbf{y}_m}{\partial \mathbf{h}_t}$

   $\mathbf{V}, \mathbf{S}, \mathbf{R} \gets \text{SVD}(\hat{\mathbf{V}})$   \Comment{Reduced SVD}
 \EndWhile

\State \textbf{return} $\mathbf{U}, \mathbf{S}, \mathbf{V}$, 
\EndProcedure
\end{algorithmic}
\end{algorithm}

\noindent \textbf{Joint and Individual Variation Explained (JIVE) Algorithm}: For completeness we provide the iterative JIVE algorithm \cite{lock2013joint}. Given a set of $N$ matrices $\mathbf{X} = [ \mathbf{X}^{(1)} \cdots \mathbf{X}^{(N)} ]$, JIVE iteratively approximates their joint and individual components.

\begin{algorithm}[H]
\caption{Joint and Individual Variation Explained}
\label{algo:jive}
\begin{algorithmic}[1]
\Procedure{JIVE}{$\mathbf{X},r_C,r_A$} 
 \State $\mathbf{X}_{joint} \gets \big[ \mathbf{X}^{{(1)}^T} \cdots \mathbf{X}^{{(N)}^T} \big]^T $ 
 
 \While{termination criterion}
 \State $\mathbf{C} = \big[ \mathbf{C}^{{(1)}^T}, \cdots,
 \mathbf{C}^{{(N)}^T}
 \big]
 \gets \text{rank} \; r_C \; \text{SVD of} \; \mathbf{X}_{joint}$

     \For{$i = (1, \cdots, N)$}
    
        \State $\mathbf{X}^{(i)}_{indiv} \gets \mathbf{X}^{(i)} - \mathbf{C}^{(i)}$

        \State $\mathbf{A}^{(i)} \gets \;\; \text{rank} \; r_A \; \text{SVD of} \; \mathbf{X}^{(i)}_{indiv} (\mathbf{I} - \mathbf{V} \mathbf{V}^T) $ \Comment{Ensures orthogonality constraint}

        \State $\mathbf{X}^{(i)}_{joint} \gets \mathbf{X}^{(i)} - \mathbf{A}^{(i)}$
     \EndFor

    \State $\mathbf{X}_{joint} = \big[  \mathbf{X}^{{(1)}^T}_{joint} \cdots \mathbf{X}^{{(N)}^T}_{joint} \big]^T$
\EndWhile

\State \textbf{return} $\mathbf{C}, \{ \mathbf{A} \}_{i=1}^N$
\EndProcedure
\end{algorithmic}
\end{algorithm}

\subsection{Impementation Details}

Here we describe more thoroughly the experimental setup we use to produce our results in
the quantitative comparisons in the main paper, in order to ensure reproducibility. For all methods compared we use an editing strength $a=50$ along the identified semantic direction to obtain the edits. The alternative methods we benchmark our proposed method against are detailed below.

\noindent \textbf{Asyrp}  For Asyrp \cite{kwon2022diffusion} we use the author’s official code\footnote{https://github.com/kwonminki/Asyrp\_official}. Since Asyrp is a supervised approach we train each attribute with 1k samples following the recipe detailed in their paper. Asyrp uses CLIP to obtain the editing directions and requires a source caption $ y^{\text{source}}$ and a target caption $y^{\text{target}}$. The source and target captions we use to obtain the editing directions for each attribute are:
\begin{itemize}
    \item Smile: $ y^{\text{source}}=$ "\textit{face}",    $ y^{\text{target}}=$ "\textit{face with a smile}"
    \item Gaze: $ y^{\text{source}}=$ "\textit{face}",    $ y^{\text{target}}=$ "\textit{face looking left}"
    \item Red Lips: $ y^{\text{source}}=$ "\textit{face}",    $ y^{\text{target}}=$ "\textit{face with red lips}"
    \item Close Eyes: $ y^{\text{source}}=$ "\textit{face}",    $ y^{\text{target}}=$ "\textit{face with closed eyes}"
\end{itemize}

\noindent \textbf{Haas et al.} We implement the method of Haas et al. \cite{haas2023discovering}  following their paper \footnote{At the time of writing the paper the \href{https://github.com/renhaa/semantic-diffusion}{official codebase}  of Haas et al. was not public yet.}. The editing directions for Smile, Close Eyes, Red Lips, and Gaze edits were obtained from CelebA-HQ samples with index numbers \texttt{00009}, \texttt{00000}, \texttt{00042}, \texttt{00003} respectively.
% \begin{itemize}
%     \item Gaze from sample 3 using principal component 5
%     \item Smile from sample 9 using principal component 3
%     \item Close Eyes from sample 0 using principal component 4
%     \item Red Lips from sample 42 using principal component 1
% \end{itemize}

\noindent \textbf{Ours} For our method we obtain all editing directions from CelebA-HQ sample with index number \texttt{00020}.

\section{Additional Experimental Resuls}

\subsection{Ablation Study: Choice of joint and individual ranks}

\begin{figure}[H]

  \centering
    \includegraphics[width=0.55\textwidth]{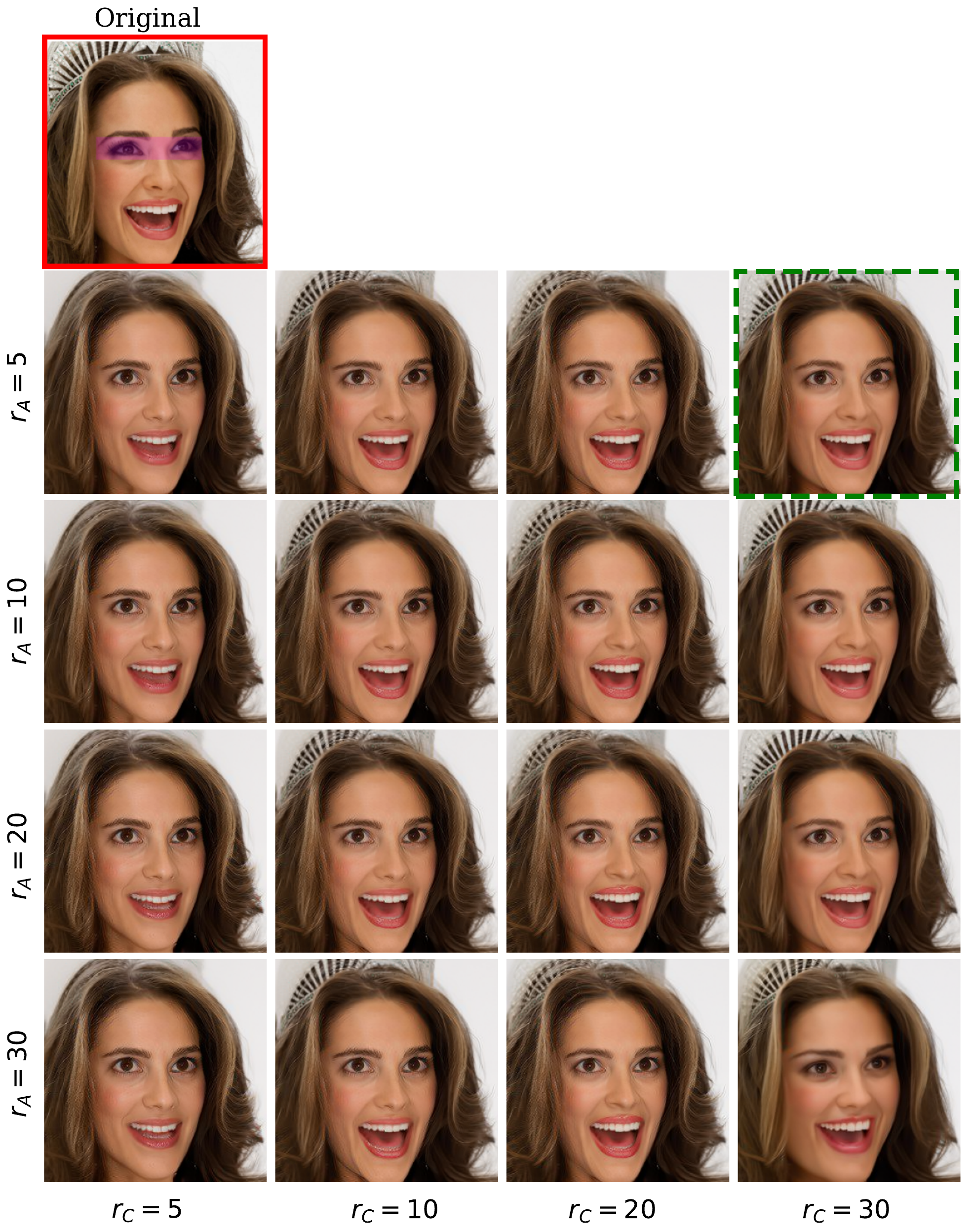}
  \vspace{5 pt}
  \caption{\textbf{Joint and individual ranks ablation}. The editing result with joint and individual ranks used for all our experiments i.e. $r_C = 30, \; r_A=5$ is highlighted with a green rectangle.}
\label{fig:rank_ablation}

\end{figure}

Here we turn our focus on the effect of joint and individual rank selection on local editing. In Figure \ref{fig:rank_ablation} we present the editing results for a localized attribute manipulation (open eyes) under various joint and individual rank choices. We highlight with a green rectangle the editing result obtained $r_C=30$ and $r_A=5$ which are the ranks used for all experiments. We observe that a low joint rank results in undesirable edits outside the region of interest, like changing the mouth and the hairband. Similar effects are observed under a high individual rank. Conversely, a low individual rank and a high joint rank produce a localized manipulation that minimally affects the rest of the image.

\subsection{Edits from different timesteps}

\begin{figure}[H]

  \centering
    \includegraphics[width=0.9\textwidth]{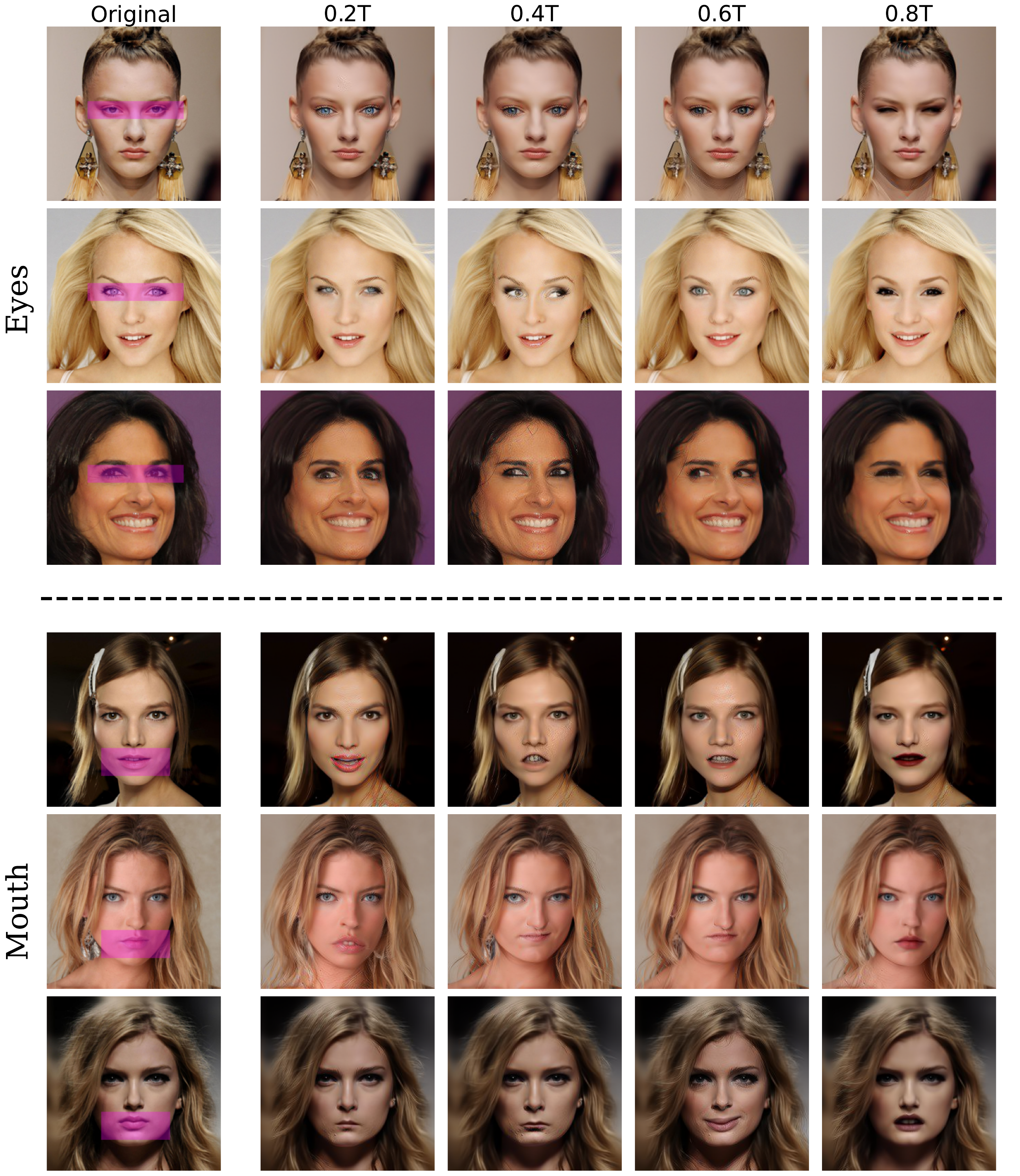}
  \vspace{5 pt}
  \caption{\textbf{Editing results from different timesteps} of the denoising process for the CelebA-HQ datasets Regions of interest are denoted by a pink rectangle. All edits presented correspond to the $1^{\text{st}}$ principal component for each timestep.}
\label{fig:timestep_celeba}

\end{figure}

\begin{figure}[H]
    \centering
    \includegraphics[width=0.95\textwidth]{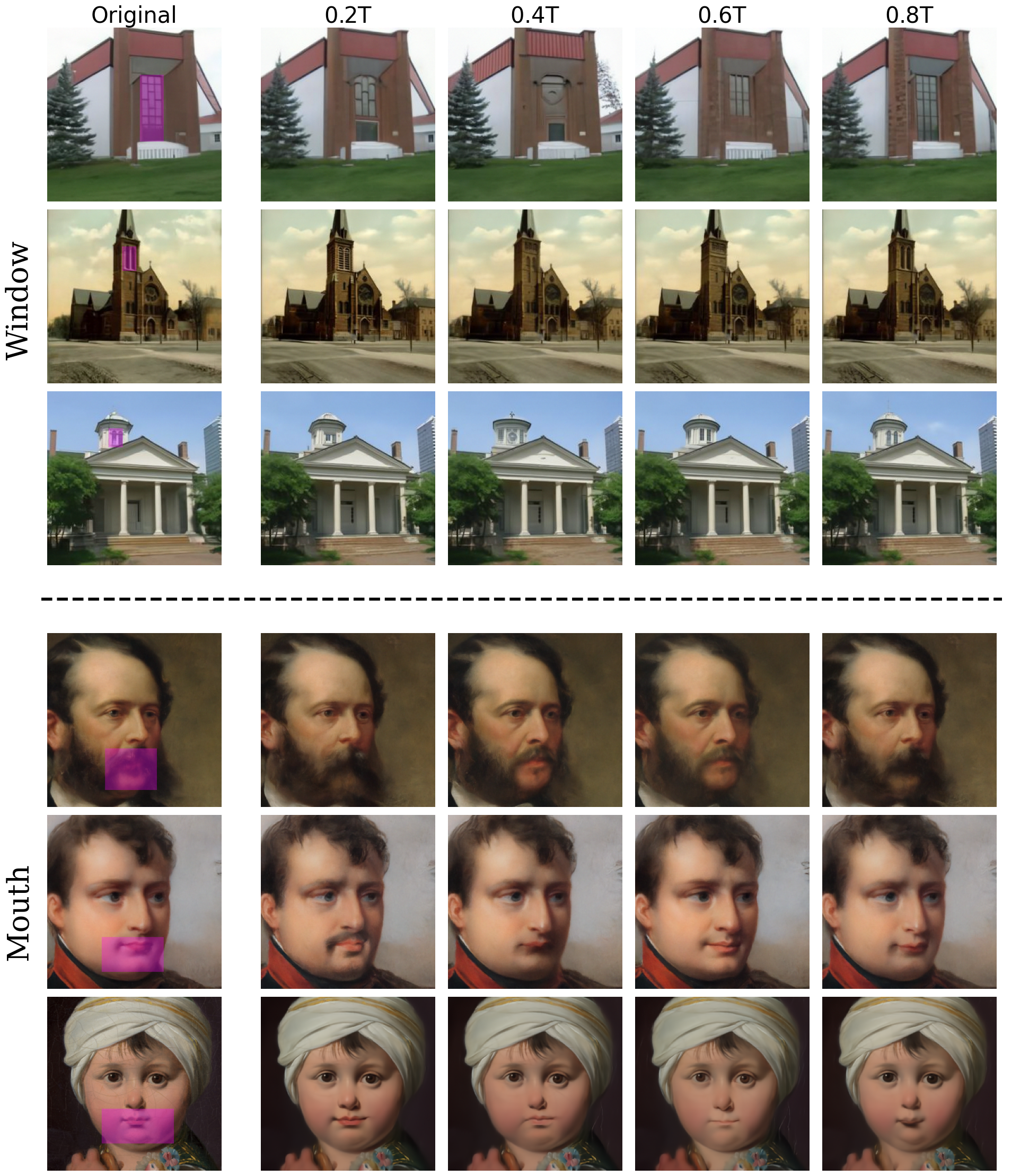}
    \vspace{5 pt}
    \caption{\textbf{Editing results from different timesteps} of the denoising process for the LSUN-Churches and M\textsc{et}F\textsc{aces} datasets. Regions of interest are denoted by a pink rectangle. All edits presented correspond to the $1^{\text{st}}$ principal component for each timestep.}
    \label{fig:timestep_other}
\end{figure}

In the main paper, we present editing results obtained at timestep $t=0.6\text{T}$ of the denoising process. In Figure \ref{fig:timestep_celeba} and Figure \ref{fig:timestep_other} we present editing results obtained from various timesteps on CelebA-HQ, LSUN-Churches, and M\textsc{et}F\textsc{aces}. All edits presented correspond to the first principal component for each timestep. We observe that our method is robust with respect to timestep selection, identifying meaningful semantic directions at various timesteps. For CelebA-HQ, these directions produce edits like changing the eye color or the gaze for the eyes region and changing the lip color and the expression for the mouth region. For LSUN-Churches, editing the window region produces different window variations, while for M\textsc{et}F\textsc{aces}, editing the mouth region changes the subject's expression and facial hair.

\subsection{Linear Interpolation between directions}

\begin{figure}[H]

  \centering
    \includegraphics[width=0.90\textwidth]{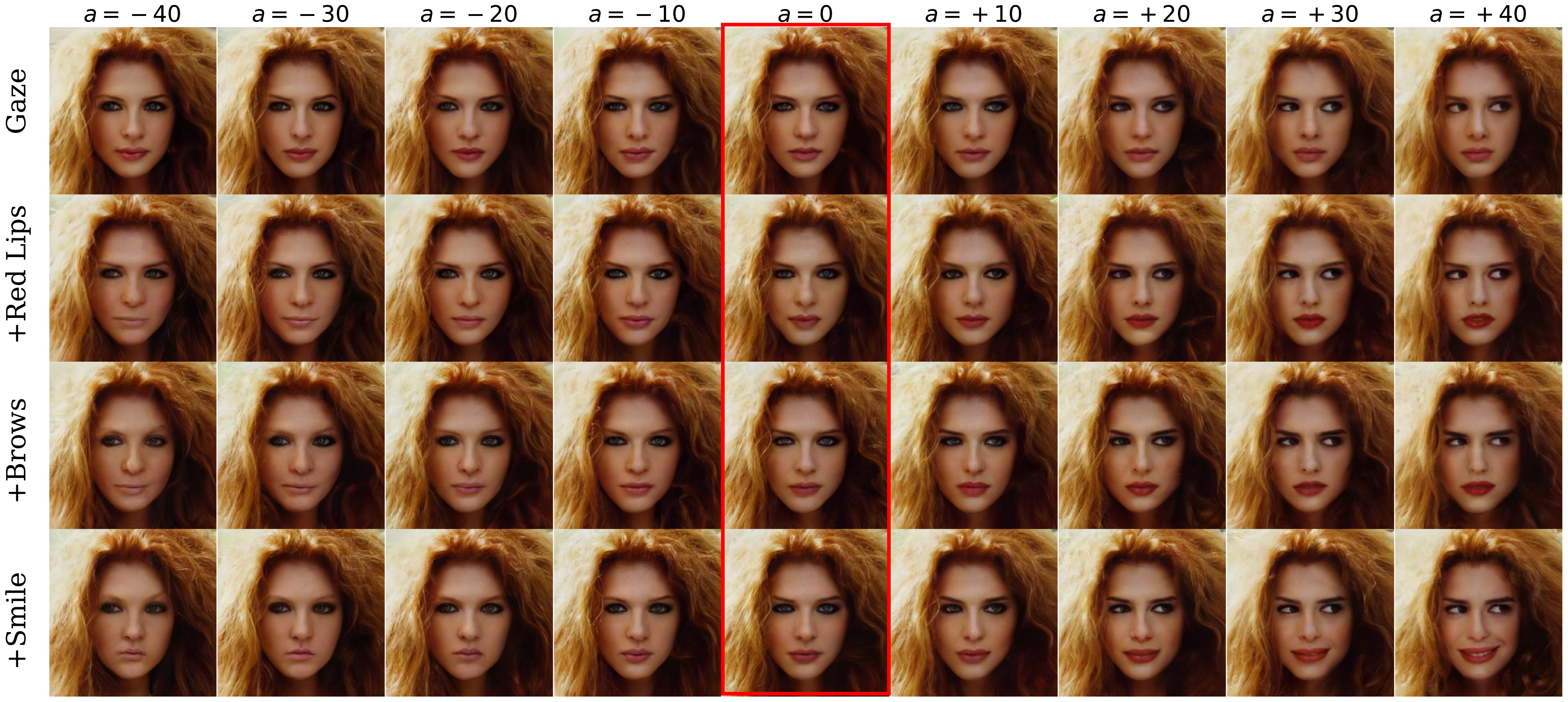}
  \vspace{5 pt}
  \caption{\textbf{Linear Interpolation between directions} for a sample from CelebA-HQ.}
\label{fig:linear}

\end{figure}

In Figure \ref{fig:linear}  we showcase how latent directions identified by our method can be composed to simultaneously edit different attributes. Let $\mathbf{v}_1, \mathbf{v}_2, \mathbf{v}_3, \mathbf{v}_4$ be the latent directions corresponding to \textit{Gaze}, \textit{Red Lips}, \textit{Brows}, and \textit{Smile} edits respectively. Each image in position $(i,j)$ in Figure \ref{fig:linear} is produced with editing vector $\mathbf{v}_{i,j}= a_i \sum_{k=1}^j \mathbf{v}_k$ with  $a_i \in [-40, ..., 40]$ and $ \mathbf{v}_k \in [\mathbf{v}_1, \mathbf{v}_2, \mathbf{v}_3, \mathbf{v}_4]$, where $a_i$ is the editing strength.
 We observe that scaling the strength of the edit controls the magnitude of attribute change and that the negative scales result in semantically opposite edits.

\subsection{Additional Comparisons}

\begin{figure}[H]

  \centering
    \includegraphics[width=0.9\textwidth]{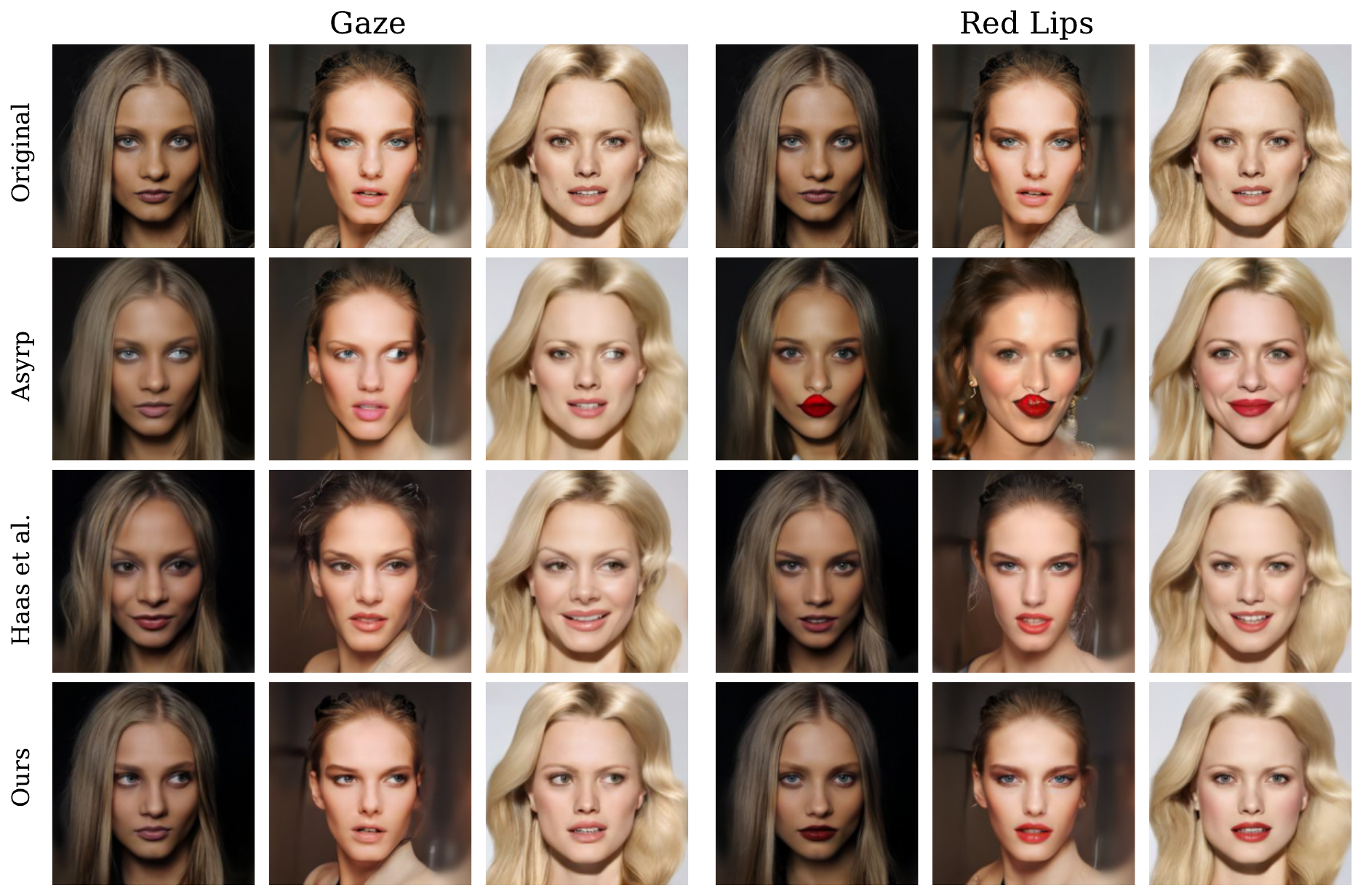}
  \vspace{5 pt}
  \caption{\textbf{Additional qualitative comparisons} between our method and existing alternatives for Red Lips and Gaze edits.}
\label{fig:compare}

\end{figure}

In Figure \ref{fig:compare} we present additional qualitative comparisons for two semantic directions not shown in the main paper. Asyrp \cite{kwon2022diffusion} produces considerable distortions and artifacts in the edited regions and in some instances significantly alters the subject's identity, as in the second and third images of the Red Lips edit. Haas et al. \cite{haas2023discovering} are better at preserving the subject's identity but the edits fail to be localized to the region of interest, changing the subject's overall expression. Our method succeeds in producing the desired semantic edits in the region of interest with almost imperceptible changes to the other parts of the image.

\bibliography{egbib}
\end{document}